\documentclass[10pt, twocolumn, letterpaper]{article}

\usepackage{cvpr}              % To produce the CAMERA-READY version
\makeatletter
\@namedef{ver@everyshi.sty}{}
\makeatother

\usepackage{graphicx}
\usepackage{amsmath}
\usepackage{amssymb}
\usepackage{booktabs}
\usepackage{breqn}
\usepackage{subcaption}
\usepackage{tikz}
\usepackage{pgfplots}
\usepgfplotslibrary[groupplots]
\usepackage{makecell}

\usepackage{caption}
\usepackage{multirow}

\usepackage{booktabs}
\usepackage{bm}
\usepackage{algorithm, algpseudocode}

\usepackage{color}
\usepackage{colortbl}
\usepackage{textcomp}
\usepackage[outercaption]{sidecap}
\usepackage{makecell}
\usepackage{stmaryrd}
\usepackage{wrapfig}
\usepackage{caption}
\usepackage{enumitem}
\usepackage{array}
\usepackage{enumitem}
\usepackage{pifont}
\usepackage{sidecap}
\usepackage{xspace}
\usepackage[export]{adjustbox}

\newcommand{\txt}[1]{{\texttt{#1}}}

\newcommand{\cmark}{\ding{51}}%
\newcommand{\xmark}{\ding{55}}%

\definecolor{Gray}{gray}{0.90}
\definecolor{LightCyan}{rgb}{0.82,0.82,1}

\usepackage[pagebackref,breaklinks,colorlinks]{hyperref}

\newcommand\given[1][]{\:#1\vert\:}

\pgfplotsset{compat=1.17}
\usepackage[capitalize]{cleveref}
\crefname{section}{Sec.}{Secs.}
\Crefname{section}{Section}{Sections}
\Crefname{table}{Table}{Tables}
\crefname{table}{Tab.}{Tabs.}

\def\cvprPaperID{9835} % *** Enter the CVPR Paper ID here
\def\confName{CVPR}
\def\confYear{2023}

\begin{document}

\title{Vita-CLIP: Video and text adaptive CLIP via Multimodal Prompting}

\author{
  Syed Talal Wasim$^{1}$ \quad 
  Muzammal Naseer$^{1}$ \quad 
  Salman Khan$^{1,2}$ \quad \\
  Fahad Shahbaz Khan$^{1,3}$
  Mubarak Shah$^{4}$
  \vspace{0.2em} \\
  $^{1}$Mohamed bin Zayed University of AI $^{2}$Australian National University \\
  $^{3}$Link\"{o}ping University $^{4}$University of Central Florida
}
\maketitle

%%%%%%%%% ABSTRACT
\begin{abstract}
    Adopting contrastive image-text pretrained models like CLIP towards video classification has gained attention due to its cost-effectiveness and competitive performance. However, recent works in this area face a trade-off. Finetuning the pretrained model to achieve strong supervised performance results in low zero-shot generalization. Similarly, freezing the backbone to retain zero-shot capability causes significant drop in supervised accuracy. Because of this, recent works in literature typically train separate models for supervised and zero-shot action recognition.
    In this work, we propose a multimodal prompt learning scheme that works to balance the supervised and zero-shot performance under a single \emph{unified} training. Our prompting approach on the vision side caters for three aspects: 1) Global \emph{video-level} prompts to model the data distribution; 2) Local \emph{frame-level} prompts to provide per-frame discriminative conditioning; and 3) a \emph{summary prompt} to extract a condensed video representation. Additionally, we define a prompting scheme on the text side to augment the textual context.
    Through this prompting scheme, we can achieve state-of-the-art zero-shot performance on Kinetics-600, HMDB51 and UCF101 while remaining competitive in the supervised setting. By keeping the pretrained backbone frozen, we optimize a much lower number of parameters and retain the existing general representation which helps achieve the strong zero-shot performance. Our codes/models are released at \href{https://github.com/TalalWasim/Vita-CLIP}{https://github.com/TalalWasim/Vita-CLIP.}.
\end{abstract}

%%%%%%%%% BODY TEXT
\section{Introduction}
\label{sec:intro}

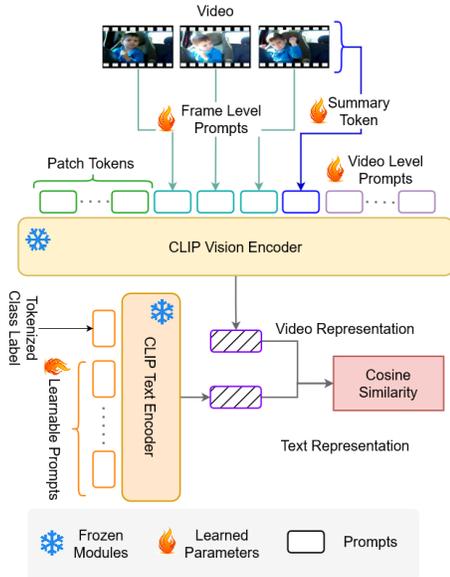
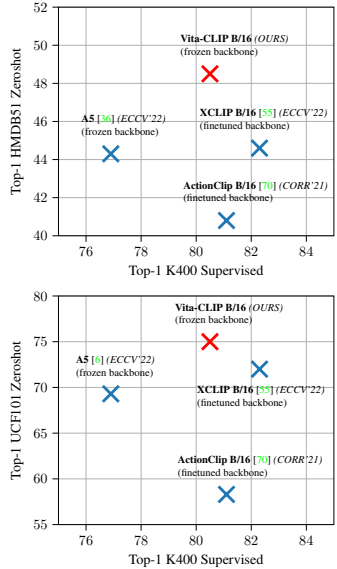
\begin{SCfigure*}
\centering
    \begin{subfigure}[t]{0.34\textwidth}
    \centering
    \raisebox{-0.5\height}{\includegraphics[width=\linewidth]{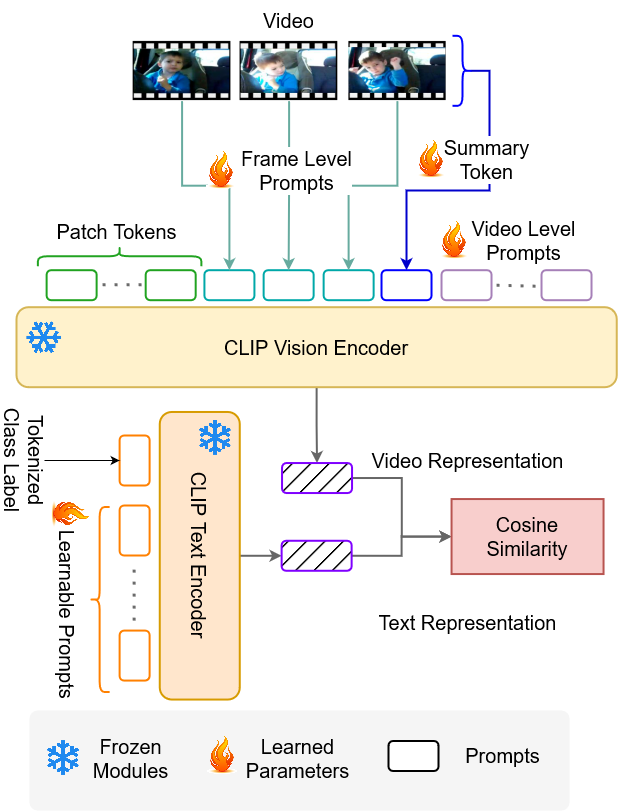}}
    \caption{Proposed Prompting Scheme}
    \end{subfigure}
    \begin{subfigure}[t]{0.3\textwidth}
    \centering\resizebox{0.85\columnwidth}{!}{
    \raisebox{-0.5\height}{\begin{tikzpicture}[every mark/.append style={mark size=8pt, line width=2pt}]

\definecolor{darkgray176}{RGB}{176,176,176}
\definecolor{steelblue31119180}{RGB}{31,119,180}

\begin{groupplot}[group style={group size=1 by 2, vertical sep=1.5cm}]
\nextgroupplot[
tick align=outside,
tick pos=left,
x grid style={darkgray176},
xlabel={Top-1 K400 Supervised},
xmajorgrids,
xmin=75, xmax=85,
xtick style={color=black},
y grid style={darkgray176},
ylabel={Top-1 HMDB51 Zeroshot},
ymajorgrids,
ymin=40, ymax=52,
ytick style={color=black}
]
\addplot [draw=steelblue31119180, fill=steelblue31119180, mark=x, only marks]
table{%
x  y
81.1 40.8
82.3 44.6
76.9 44.3
};
\addplot [draw=red, fill=red, mark=x, only marks]
table{%
x  y
80.5 48.5
};
\draw (axis cs:79.3,49.5) node[
  scale=0.75,
  anchor=base west,
  text=black,
  rotate=0.0,
  align=left
]{\textbf{Vita-CLIP B/16} \textit{(OURS)}\\
(frozen backbone)};
\draw (axis cs:79.4,41.8) node[
  scale=0.75,
  anchor=base west,
  text=black,
  rotate=0.0,
  align=left
]{\textbf{ActionClip B/16}\cite{action-clip} \textit{(CORR'21)}\\
(finetuned backbone)};
\draw (axis cs:80.0,45.6) node[
  scale=0.75,
  anchor=base west,
  text=black,
  rotate=0.0,
  align=left
]{\textbf{XCLIP B/16}\cite{xclip} \textit{(ECCV'22)}\\
(finetuned backbone)};
\draw (axis cs:75.7,45.3) node[
  scale=0.75,
  anchor=base west,
  text=black,
  rotate=0.0,
  align=left
]{\textbf{A5}\cite{video-coop} \textit{(ECCV'22)}\\
(frozen backbone)};

\nextgroupplot[
tick align=outside,
tick pos=left,
x grid style={darkgray176},
xlabel={Top-1 K400 Supervised},
xmajorgrids,
xmin=75, xmax=85,
xtick style={color=black},
y grid style={darkgray176},
ylabel={Top-1 UCF101 Zeroshot},
ymajorgrids,
ymin=55, ymax=80,
ytick style={color=black}
]
\addplot [draw=steelblue31119180, fill=steelblue31119180, mark=x, only marks]
table{%
x  y
81.1 58.3
82.3 72
76.9 69.3
};
\addplot [draw=red, fill=red, mark=x, only marks]
table{%
x  y
80.5 75
};
\draw (axis cs:79.1,77) node[
  scale=0.75,
  anchor=base west,
  text=black,
  rotate=0.0,
  align=left
]{\textbf{Vita-CLIP B/16} \textit{(OURS)}\\
(frozen backbone)};
\draw (axis cs:79.2,60.3) node[
  scale=0.75,
  anchor=base west,
  text=black,
  rotate=0.0,
  align=left
]{\textbf{ActionClip B/16}\cite{action-clip} \textit{(CORR'21)}\\
(finetuned backbone)};
\draw (axis cs:79.9,68) node[
  scale=0.75,
  anchor=base west,
  text=black,
  rotate=0.0,
  align=left
]{\textbf{XCLIP B/16}\cite{xclip} \textit{(ECCV'22)}\\
(finetuned backbone)};
\draw (axis cs:75.5,71.3) node[
  scale=0.75,
  anchor=base west,
  text=black,
  rotate=0.0,
  align=left
]{\textbf{A5}\cite{timesformer2021} \textit{(ECCV'22)}\\
(frozen backbone)};
\end{groupplot}

\end{tikzpicture}}}
    \caption{Zero-shot accuracy (HMDB51, UCF101) vs supervised accuracy (Kinetics-400).}
    \end{subfigure}
    \caption{An \textbf{overview of the proposed prompting scheme} (\emph{left}) alongside the trade-off which we attempt to balance between supervised and zero-shot performance (\emph{right}). \textbf{(a)} Our prompting approach adds learnable parameters to learn {visual} and temporal information in videos at three levels: a \emph{summary prompt} to learn a condensed representation of the video, \emph{video-level prompts} to model global distribution shifts needed to adapt to video domain and \emph{frame-level prompts} to enrich local discriminative information in each frame. On the text side, we learn prompts to adapt the language representations for videos. \textbf{(b)} The trade-off plots showing zero-shot vs. supervised  performance comparison for ours and recent CLIP-based video approaches. Note that existing SoTA \cite{xclip} trains two separate models for zero-shot and supervised settings while our method offers a unified model with the same training for both settings.}
    \label{fig:intro}
\end{SCfigure*}

In the image classification domain, multimodal image-text pretrained models such as CLIP \cite{clip}, ALIGN \cite{align} and Florence \cite{florence} have shown the capability of learning generalized representations. These models, trained on large-scale language-image pairs in a contrastive manner, have remarkable zero-shot capabilities and transfer well to a variety of downstream tasks. However, training a similar model for the task of video recognition is not feasible both in terms of gathering large-scale video-text pairs, which can suffer from alignment problems \cite{alignment}, and is also exponentially more computationally expensive due to multiple frames being processed per video. Therefore, there has been a recent push in the research community to effectively adopt the pretrained image-text models for the task of video recognition, while maintaining their zero-shot capabilities.
In this regard, existing methods can be divided into two categories. Some take inspiration from recent prompt learning methods \cite{coop,cocoop,clip-adapter,tip-adapter,vpt} and propose a prompt learning scheme either on the text \cite{video-coop} or vision \cite{xclip,action-clip} side, along with additional transformer layers for improved temporal learning. Others prefer an end-to-end CLIP finetuning scheme for video tasks \cite{xclip,clip4clip,action-clip}. However, the problem with these methods is that they either fail to effectively leverage learning on both the text and vision sides \cite{xclip,video-coop} or end up losing the zero-shot generalization of CLIP by finetuning the vision  decoder \cite{frozen-clip} or the backbone \cite{action-clip,xclip,clip4clip}. In summary, the existing approaches can steer the model \emph{either} towards good zero-shot generalization \emph{or} better supervised learning on video tasks. Since real-world tasks require both supervised and zero-shot capabilities, our work investigates the following question: \emph{Can we develop a \emph{unified} model for videos that performs well for both supervised learning and zero-shot generalization tasks?}

In pursuit of the aforementioned question, we propose a multimodal prompting-based Video and text adaptive CLIP. To effectively adapt the pretrained image-text CLIP model to videos, we consider two important aspects. Firstly, one needs to preserve the generalization capabilities of the original pretrained CLIP backbone and secondly, it must be able to effectively adapt to the video domain. In this regard, we propose to keep the entire backbone frozen and learn additional lightweight modules to adapt the model for videos. On this point, for the vision side, we aim to explicitly exploit the temporal information in videos which is lacking in the frozen image model. Our approach models video information at three levels: first via \emph{global video-level prompts} that learn the overall distribution characteristics of video data \eg, motion and dynamics; secondly, inspired by \cite{naseer2023boosting}, \emph{local frame-level prompts} which model per frame discriminative information by directly conditioning on classification tokens of all frames; and thirdly by a \emph{summary prompt} that distills the entire video sequence response in a single concise summary vector. 

Additionally, to better model the textual context we propose to use a learnable context on the text encoder. The reason why this is particularly important is that the textual information is quite limited in the available video datasets. Instead of having per-sample text descriptions, we are limited to using class labels as text descriptions. 
%Thus, we propose to use a learning scheme on the text side to augment the video class label descriptions. 
Inspired by \cite{coop}, we propose a prompt learning method on the text side to better model the textual context and to augment the video class label descriptions. An overview of our method with the trade-off it seeks to balance is presented in \autoref{fig:intro}. The main contributions of this work are as follows:
\begin{itemize}[leftmargin=*]\setlength{\itemsep}{0em}
    \item We propose a multimodal prompting approach Vita-CLIP for videos that learns video and text-specific context vectors to efficiently adapt the image-text pretrained CLIP model to video recognition tasks.
    \item On the vision side, we explicitly model the temporal information and the video data distribution. Our prompt learning method aggregates the discriminative information from each frame in a clip with every other frame, while also providing per-layer learning capacity to better capture the data distribution. On the language side, our approach learns complimentary semantic context to better adapt the language representations.
    \item We evaluate our approach on supervised as well as generalization tasks and demonstrate a sound balance between both aspects using a \emph{single} unified model. Specifically, on zero-shot tasks, we obtain 4.0\%, 3.0\% and 2.2\% gains over the recent SoTA X-CLIP \cite{xclip} on HMDB-51, UCF-101, and Kinetics-600 datasets respectively. 
\end{itemize}

\section{Related work}
\textbf{Vision-Language (VL) Models:} VL models \cite{clip, align, florence} consists of an image and a text encoder and are trained on large-scale image-text pairs in a contrastive manner to learn a common feature space between images and textual labels. The semantic supervision driven by text allows models like CLIP \cite{clip} to learn fine-grained visual concepts which are transferable to many downstream tasks; semantic segmentation \cite{rao2022denseclip, ghiasi2021open, zhou2021denseclip}, object detection \cite{du2022learning}, point cloud classification \cite{zhang2022pointclip}, and video classification \cite{xu2021videoclip}. Importantly, these models allow `zero-shot' knowledge transfer. In the video domain, there exist some models trained with video-text pairs for applications such as video retrieval \cite{anne2017localizing, lei2021less, miech2020end}. However, these models are not trained on large amounts of video-text data. In this work, we propose a novel approach to induce temporal cues within the pretrained VL model, CLIP, to enhance its `zero-shot' generalization on videos.

\textbf{Video Recognition:} The conventional techniques for spatiotemporal learning for video recognition progressed from hand-crafted features \cite{dollar2005behavior, klaser2008spatio, wang2013dense}  to end-to-end deep learning methods \cite{lecun2015deep}. Among neural network-based approaches, 3D convolutional networks (CNNs) \cite{carreira2017quo, diba2018spatio, feichtenhofer2019slowfast, stroud2020d3d} learn spatiotemporal representation directly from RGB video data, while other methods deploy dedicated 2D CNNs \cite{feichtenhofer2016convolutional, wang2016temporal, wang2017spatiotemporal, jiang2019stm} and learn spatial and dynamic information within separate networks before fusing them together. The trade-off between 2D/3D networks for videos has been explored in \cite{tran2018closer, xie2018rethinking, zhou2018mict}. Recently, Transformer \cite{dosovitskiy2020image} based architectures have emerged for video recognition \cite{arnab2021vivit, bertasius2021space, neimark2021video, fan2021multiscale, ranasinghe2022selfsupervised}. In this work, we propose to adopt a pretrained multi-modal Transformer \cite{clip} for spatiotemporal learning.

\textbf{Prompt Learning:} Prompting was proposed in NLP domain \cite{liu2021pre, jiang2020can} and it refers to generating task-specific instructions to get the desired behavior from language models. These instructions can be created manually \cite{brown2020language} or learned by training discrete \cite{gao2020making, jiang2020can, rohrbach2017movie, schick2020exploiting} or continuous vectors \cite{lester2021power, li2021prefix}. Prompt learning has recently been explored in vision problems to transfer knowledge from large-scale models to downstream tasks. The current prompting techniques are applied to both uni-models \eg, ViTs trained on images \cite{dosovitskiy2020image} as well as multimodal models such as CLIP. For the case of ViTs, \cite{jia2022visual, bahng2022visual} train learnable prompts to steer pretrained vision transformers \cite{dosovitskiy2020image, liu2021swin}. On the other hand, methods like \cite{coop, cocoop, sun2022dualcoop} introduce learnable vectors into the text encoder of CLIP for transfer learning to image recognition tasks. In contrast, we propose to learn multimodal video prompts to steer both vision and text encoders of CLIP simultaneously for spatiotemporal learning on videos.

\textbf{Adapting VL Models for Videos:} CLIP model has been fully fine-tuned on video-based retrieval and recognition tasks \cite{clip4clip, action-clip}. Ju \etal \cite{video-coop}  transfer the zero-shot generalization capability of CLIP to videos by learning prompts on the text encoder inputs and two transformer layers on the frame-level visual representations from the image encoder to model temporal context. However, directly using the CLIP image encoder for videos leads to a lack of temporal information within earlier blocks of the CLIP vision encoder and as a consequence, such an approach shows less generalization than full-fine tuning \cite{action-clip}. Similarly, \cite{xclip} proposes a cross-frame attention module to model long-range inter-frame dependencies in videos and uses text prompt generation conditioned on video and text representations for better generalization. In contrast to these methods, we introduce a learnable video prompting module within the image and text encoder of CLIP to model temporal cues without full fine-tuning and demonstrate a good trade-off between generalization and fully supervised performance.

\section{Vita-CLIP: Methodology}

Our approach, Vita-CLIP, works to adapt pretrained image-based vision-language models for videos using a multimodal prompting scheme that aims to retain both the strong generalization capability (zero-shot performance) as well as good supervised performance. Vita-CLIP allows utilizing the existing image-language pretrained model rather than training one from scratch for videos.

This section presents our approach. We start with an overview of the vision/text encoders in \autoref{sec:meth:ovr_VT}, followed by a detailed explanation of our multimodal prompt learning scheme in \autoref{sec:meth:mmpl}. This is further divided into vision (\autoref{sec:meth:vepl}) and text-side prompt learning (\autoref{sec:meth:tepl}). Finally, we outline our learning objective in \autoref{sec:meth:ovr_LO}.

\begin{SCfigure*}[][t]
    \centering
    \includegraphics[width=.7\textwidth]{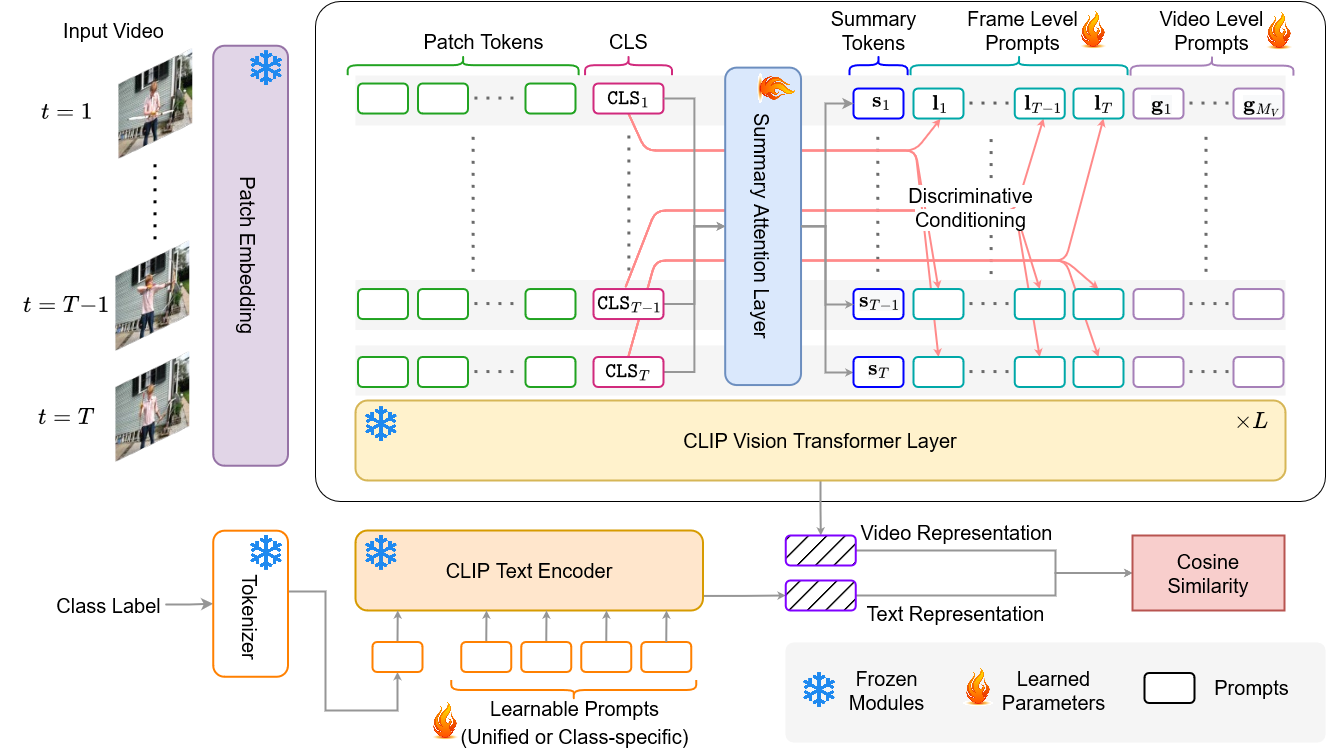}
    \caption{\textbf{Vita-CLIP Prompting Architecture:} We append multiple prompt tokens both on the vision and text encoders. On the vision encoder, we infer a Summary Token ($S$) which condenses the whole video token sequence which is appended with the input. Additionally, we add $M_v$ number of Global ($G$) video-level prompts to learn the data distribution and ($T$) number of frame-level prompts conditioned on the respective frame's $\texttt{CLS}$ token to reinforce discriminative information. On the text side, we add $M_c$ number of learnable prompts to model the input context of the text encoder. Modules with (\protect\includegraphics[height=0.3cm]{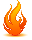}) are trainable and those with (\protect\includegraphics[height=0.3cm]{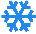}) are frozen.}
    \label{fig:method}
\end{SCfigure*}

\subsection{Video and Text Encoding}
\label{sec:meth:ovr_VT}
As stated earlier, we wish to adopt the pretrained image-text models to videos in a manner that we retain both the pretrained generalized representation, while also achieving competitive fully-supervised performance with methods that employ finetuning on the text and/or vision encoders. In that regard, we propose a multimodal vision and text prompt learning scheme that keeps both the vanilla CLIP image and text encoders frozen and introduces extra learnable parameters to adopt them for videos. From a broader perspective, we obtain video (${\bf{v}}$) and text (${\bf{c}}$) representations from the video ($f_{\theta_v}$) and text ($f_{\theta_c}$) encoders respectively. This section formally defines how these representations are obtained, while specific details on the proposed prompt learning scheme are presented in \autoref{sec:meth:mmpl}.

\textbf{Video Encoder:}
Consider a video $V \in \mathbb{R}^{T\times H \times W \times 3}$ of spatial size $H \times W$ with $T$ sampled frames. Each frame $t \in \{1 \ldots T\}$ is divided into $N$ non-overlapping square patches of size $P\times P$ as required by the ViT architecture \cite{dosovitskiy2020image}, with the total number of patches being $N = H\times W/P^2$. 
For each frame, all patches of shape $P\times P \times 3$ are flattened into a set of vectors and represented as $\{{\bf{x}}_{t,i} \in \mathbb{R}^{3P^2} \}_{i=1}^{N}$, where $t$ is the frame number and $i$ the patch number. 
The vectors are then projected to form token embeddings using a linear projection layer ${\mathbf{P}_{emb}} \in \mathbb{R}^{3P^2 \times D}$, with an output dimension $D$ for each token. 
An additional classification token, ${\bf{x}}_{cls} \in \mathbb{R}^D$, is prepended to the sequence of embedded tokens for each frame. 
The final per-frame token sequence fed into the video encoder is given by:
\begin{equation}
{\bf{z}}^{(0)}_{t} = [{\bf{x}}_{cls}, {\mathbf{P}}_{emb}^T \ \mathbf{x}_{t,1}, \cdots, {\mathbf{P}}_{emb}^T \ \mathbf{x}_{t,N}] + {\bf{e}}, 
\end{equation}
where ${\bf{e}} = {\bf{e}}^{sp} + {\bf{e}}^{tm}$. Here, ${\bf{e}}^{sp}$ and ${\bf{e}}^{tm}$ denote the spatial and temporal positional encodings, respectively. %, that are added to the input sequence of tokens.

From the $L_v$ layered video encoder, we obtain the frame level representation at each layer $l$ as follows:
\begin{equation}
    {\bf{z}}^{(l)}_{t} = f_{\theta_v}^{(l)}({{\bf{z}}^{(l-1)}_{t}}), \quad l \in \{ 1, \cdots, L_v\}, 
\end{equation}
where $f_{\theta_v}^{(l)}$ is the $l$-th layer of the video encoder.

Finally, to obtain the per-frame representation, the classification token $\mathbf{x}_{cls}$ is extracted from the output token sequence of the last layer ${\bf{z}}^{(L_v)}_{t}$, and projected to a dimension $D'$ using a linear projection layer ${\bf{P}}_{out} \in \mathbb{R}^{D \times D'}$.
\begin{equation}
{\bf{v}}_t = {\bf{P}}_{out}^T \ {\bf{z}}^{(L_v)}_{t,0} \; \in \mathbb{R}^{D'},
\end{equation}
where ${\bf{v}}_t$ is the output representation of frame $t$ and ${\bf{z}}^{(L_v)}_{t,0}$ is the classification token from the output sequence of the last layer of the video encoder. To obtain the video representation, the per-frame representations ${\bf{v}}_t$ are simply average-pooled to obtain the aggregate representation:
\begin{equation}
    {\bf{v}} = \operatorname{AvgPool}([{\bf{v}}_1, \cdots, {\bf{v}}_T]).
    % \vspace{-.1cm}
\end{equation}

\textbf{Text Encoder:}
For the input text representation, a pretrained text encoder is used with an additional text prompt learning scheme. 
%Specific details of the prompt learning scheme are presented in \autoref{sec:meth:mmpl}.
The pretrained text encoder is a $12$ layer BERT \cite{bert} model (for CLIP B/16 variant) with an embedding size of $512$ and context length of $77$. Each layer of the model consists of a Multi-Head Self Attention (MHSA) followed by a Feed-Forward Network (FFN). Given the text description $C$ for a video, we use the text encoder to obtain a representation ${\bf{c}} = f_{\theta_c}(C)$. Rather than using a hand-crafted prompt for the text description like ``\txt{A video of the action of \{label\}}", as used in recent works \cite{action-clip}, we use a prompt learning scheme inspired by recent works on text prompting for language-image models \cite{coop, cocoop}. %Further details on the prompt learning scheme are presented in \autoref{sec:meth:mmpl}.

%\subsection{Multi-Modal Prompt Learning}
\subsection{Video and text Prompt Learning}
\label{sec:meth:mmpl}

While there have been previous attempts at prompt learning to adapt language-image models to videos, they either focus on just the vision or text sides \cite{xclip, video-coop} coupled with completely finetuning the entire vision encoder in some cases \cite{xclip, action-clip}.
To adapt our pretrained language-image model for videos, we propose a novel multimodal prompt learning scheme that keeps the pretrained model frozen, to better retain its general representation. By preserving this representation we are able to train a \emph{single} model that can perform well both in supervised and zero-shot settings, unlike recent works \cite{xclip} that require different hyper-parameter choices to produce separate models for each setting.

In that regard, our multimodal prompting aims to align the pretrained representation towards the video tasks, ensuring that both text and vision information is utilized. More specifically, on the text side, we introduce a learnable context rather than a hand-crafted prompt to allow for the text encoder to better adapt to the new video categories. On the vision side, we propose a video prompting scheme that focuses on modeling the frame-level information and inter-frame relationships as well as  providing adaptability to new video data distributions. We explain our video and text prompting in \autoref{sec:meth:vepl} and \autoref{sec:meth:tepl} respectively.

\subsubsection{Video Encoder Prompt Learning}
\label{sec:meth:vepl}

For prompting on the vision encoder we have two major objectives: 1) Exploiting the temporal information by introducing information exchange between frames, and 2) providing additional parameters to adapt the CLIP image representations towards the video dataset distribution.

In that regard, we introduce three kinds of additional tokens which are appended to the token sequence ${\bf{z}}^{(l)}_{t}$ from frame $t$ at layer $l$. Specifically, at each layer, we introduce a single \emph{summary token} which summarises the discriminative information across all frames, $T$ frame level \emph{local prompt tokens} to communicate per-frame discriminative information to the rest of the frames in the clip and $M_v$ video-level \emph{global prompt tokens} to provide learning capacity to adapt the model to the video dataset distribution. Detailed descriptions of these types of prompt tokens are given below.

\noindent\textbf{Summary Token:} The summary token is inspired by the concept of message attention proposed in \cite{xclip}. It is used to summarize the discriminative information from each frame in the clip and provide it back to every frame, before applying the pretrained self-attention for that layer. More specifically the summary token $\mathbf{s}^{(l)}_t$ at the $l$-th layer for the $t$-th frame is obtained by first applying a linear projection $\mathbf{P}_{sum}$ on the classification tokens ${\bf{z}}^{(l-1)}_{t,0}$ and then applying a $\operatorname{MHSA}$ operation between these frame-level tokens:
\begin{align}
        {\bf{Z}}^{(l-1)}_{0, proj} &= \mathbf{P}_{sum}^T {\bf{Z}}^{(l-1)}_{0}, \notag \\
        S^{(l)} &=  \operatorname{MHSA}({\operatorname{LN}( {\bf{Z}}^{(l-1)}_{0, proj})}) + {\mathbf{Z}}^{(l-1)}_{0, proj},
\end{align}
where $\mathbf{Z}^{(l-1)}_{0} = [{\bf{z}}^{(l-1)}_{1,0}, \cdots, {\bf{z}}^{(l-1)}_{T,0}]$, $\mathbf{S}^{(l)} = [{\bf{s}}^{(l)}_{1}, \cdots, {\bf{s}}^{(l)}_{T}]$ and $\operatorname{LN}$ indicates layer normalization. Afterward, the respective summary token is appended to the token sequence ${\bf{z}}^{(l-1)}_{t}$ before applying the frozen pretrained self-attention for that layer as indicated by \autoref{eq:fsa}.

\noindent\textbf{Global Prompt Tokens:} The video-level \emph{global prompt tokens} ($\mathbf{G}^{(l)} = [\mathbf{g}^{(l)}_{1}, \cdots, \mathbf{g}^{(l)}_{M_v}]$) are randomly initialized learnable vectors. They are used to provide the model with additional learning capacity to learn the data distribution.

\noindent\textbf{Local Prompt Tokens:} The frame-level \emph{local  prompt tokens} ($\mathbf{L}^{(l)} = [\mathbf{l}^{(l)}_{1}, \cdots, \mathbf{l}^{(l)}_{T}]$) are also randomly initialized learnable vectors, equal to the number of frames, $T$, in the clip during training, but they are conditioned on the respective classification tokens for each frame. This conditioning of $\mathbf{L}^{(l)}$ on \txt{[CLS]} token ${\mathbf{z}}^{(l-1)}_{t,0}$ enables a top-down discriminative information flow in frame-wise learnable tokens. Each frame-level \emph{local prompt token} is defined as:
\begin{align}
    \hat{\mathbf{l}}^{(l)}_{t} = {\mathbf{l}}^{(l)}_{t} + {\mathbf{z}}^{(l-1)}_{t,0}.
\end{align}

Finally, the tokens $\hat{\mathbf{L}^{(l)}} = [\hat{\mathbf{l}}^{(l)}_{1}, \cdots, \hat{\mathbf{l}}^{(l)}_{T}]$ and $\mathbf{G}^{(l)} = [\mathbf{g}_{1}^{(l)}, \cdots, \mathbf{g}_{M_v}^{(l)}]$ are appended to each frame sequence ${\bf{z}}^{(l-1)}_{t}$ before applying the frozen pretrained self-attention ($\operatorname{FSA}$) for that layer as indicated below,
\begin{align}
    \label{eq:fsa}
    [\hat{{\mathbf{z}}}^{(l)}_{t}, \mathbf{S}^{(l)}, \mathbf{G}^{(l)}, \mathbf{L}^{(l)}] = & \operatorname{FSA}(\operatorname{LN}([{\mathbf{z}}^{(l-1)}_{t}, \mathbf{S}^{(l)}, \mathbf{G}^{(l)}, \mathbf{L}^{(l)}]))  \notag\\
    & + [{\mathbf{z}}^{(l-1)}_{t}, \mathbf{S}^{(l)}, \mathbf{G}^{(l)}, \mathbf{L}^{(l)}],
\end{align}
%where  is the frozen self-attention at layer $l$ of the pretrained model. 
Finally, we remove the extra appended tokens and apply the feed-forward network ($\operatorname{FFN}$) only on $\hat{{\bf{z}}}^{(l)}_{t}$ as shown below:
\begin{equation}
    {\bf{z}}^{(l)}_t = \operatorname{FFN}(\operatorname{LN}(\hat{{\bf{z}}}^{(l)}_t)) +  \hat{{\bf{z}}}^{(l)}_t.
    % \vspace{-.1cm}
\end{equation}

\subsubsection{Text Encoder Prompt Learning}
\label{sec:meth:tepl}
Inspired from \cite{coop, cocoop, video-coop}, we also use a prompt learning scheme on the text encoder. Rather than hand-crafting a textual input based on the class labels, we model the context words using trainable vectors. More specifically the input to the text encoder, $f_{\theta_c}$, is a sequence of tokens of the form:
\begin{equation}
    C = [\mathbf{u}_1, \mathbf{u}_2 , \cdots, \mathbf{u}_{M_c}, \texttt{\{label\}}]
\end{equation}
where $\mathbf{u}_i, i \in \{1, \cdots, M_c\}$ is a trainable vector of the same size as the input embeddings of the text encoder, and $M_c$ is the number of trainable unified prompts. This token sequence is then passed to the text encoder which produces the text embedding ${\bf{c}} = f_{\theta_t}(C)$.

While two different variations are possible, Unified Context (UC) (where all classes share a single set of context vectors) and Class-Specific Context (CSC) (where an independent set of context vectors is defined for each class), we use CSC in our methodology. The prompt vectors are defined as $[\mathbf{u}_{i}^{n_c}], i \in \{1, \cdots, M_c\}$ and $n_c \in \{1, \cdots, N_c\}$ where $N_c$ is the total number of classes. The effectiveness of using CSC over UC is shown through ablations in \autoref{sec:exp:ablation}.

The class-specific prompts are used in all our experiments except the zero-shot ones, where novel classes can appear. For the case of zero-shot evaluation, we simply use manual prompts with any given class name.

\subsection{Learning Objective}
\label{sec:meth:ovr_LO}

%Conventional methods video recognition have traditionally used the common Cross-Entropy loss based training for a closed-set classification problem \cite{arnab2021vivit, bertasius2021space, neimark2021video, fan2021multiscale, carreira2017quo, diba2018spatio, feichtenhofer2019slowfast, stroud2020d3d}. However, recently there has been a trend to employ text supervision for video recognition rather than one-hot labels, inspired from the works on contrastive language-image pretraining \cite{clip, align, florence}.

As explained above, our architecture consists of a Vision Transformer (ViT) \cite{dosovitskiy2020image} based image encoder and a BERT \cite{bert} text encoder similar to CLIP \cite{clip}. The vision and text encoders encode the video and text descriptions respectively, which are then compared using a cosine similarity objective. More formally, given a set of videos $\mathcal{V}$ and a set of text class descriptions $\mathcal{C}$, we sample video $V \in \mathcal{V}$ and an associated text description $C \in \mathcal{C}$ which are then passed to the video ($f_{\theta_v}$) and text ($f_{\theta_c}$) encoders respectively. This results in the video and text representations are given as:
\begin{equation}
\label{eq:rep}
{\bf{v}} = f_{\theta_v}(V \given \mathbf{S}^{(l)}, \mathbf{G}^{(l)}, \mathbf{L}^{(l)} ), \\  {\bf{c}} = f_{\theta_t}(C).
\end{equation}

We then define the cosine similarity loss function $\mathcal{L}_{cos}$ between the video and text representations as below:
\begin{equation}
    \mathcal{L}_{cos}({\bf{v}}, {\bf{c}}) = \frac{\langle {\bf{v}}, \bf{c}\rangle}{\left\|{\bf{v}}\right\| \left\|\bf{c}\right\|}.\vspace{-.15cm}
\end{equation}
We aim to maximize $\mathcal{L}_{cos}$ for the true ${\bf{v}}$ and ${\bf{c}}$ pairs and minimize otherwise.

\section{Results and Analysis}
\label{sec:exp}

% In this section we present our experimental settings, outlining procedure for conducting experiments in different settings like supervised and zero-shot. We follow this with detailed ablations for our methodology.

% tables: supervised
\begin{table*}[ht]

\centering
\small
\caption{Comparison with state-of-the-art on Kinetics-400 \cite{k400} Supervised Training. We compare with various initializations (Random, ImageNet 1k/21k, and CLIP-400M), specifying the number of frames, views, and FLOPs. We also mention whether the models use a frozen/fine-tuned backbone and whether the method is suitable for zero-shot evaluation.}
\setlength{\tabcolsep}{3mm}
\scalebox{0.85}[0.85]{
\begin{tabular}{lccccccccc}
\rowcolor{Gray}
    \toprule
    Method                                                           & Pre-training  & Finetuning & Frames   & Views     & Top-1 & Top-5 & GFLOPs    & Zero-shot \\
    \midrule
    \multicolumn{1}{l}{\textit{Initialization: Random weights}} \\
    MViTv1-B, 64×3~\textit{(ICCV'21)}~\cite{Fan_2021_ICCV}           & \xmark        & \cmark    & 64        & 3 × 3     & 81.2  & 95.1  & 455       & \xmark \\
    \midrule
    \multicolumn{7}{l}{\textit{Initialization: ImageNet weights}} \\
    Uniformer-B~\textit{(ICLR'22)}~\cite{li2022uniformer}            & IN-1k         & \cmark    & 32        & 4 × 3     & 83.0  & 95.4  & 259       & \xmark \\ 
    TimeSformer~\textit{(ICML'21)}~\cite{timesformer2021}            & IN-21k        & \cmark    & 96        & 1 × 3     & 78.0  & 93.7  & 590       & \xmark \\
    Mformer~\textit{(NeurIPS'21)}~\cite{patrick2021keeping}          & IN-21k        & \cmark    & 16        & 10 × 3    & 79.7  & 94.2  & 370       & \xmark \\
    Swin-B~\textit{(CVPR'22)}~\cite{liu2021video}                    & IN-1k         & \cmark    & 32        & 4 × 3     & 80.6  & 94.6  & 282       & \xmark \\
    Swin-B~\textit{(CVPR'22)}~\cite{liu2021video}                    & IN-21k        & \cmark    & 32        & 4 × 3     & 82.7  & 95.5  & 282       & \xmark \\
    MViTv2-B~\textit{(CVPR'22)}~\cite{li2021mvitv2}                  & \xmark        & \cmark    & 32        & 5 × 1     & 82.9  & 95.7  & 225       & \xmark \\
    \midrule
    \multicolumn{7}{l}{\textit{{Initialization: Large-scale image-language weights (finetuned backbone)}}} \\
    ActionCLIP-B/16~\textit{(arXiv'21)}~\cite{action-clip}           & CLIP-400M     & \cmark    & 32        & 10 × 3    & 83.8  & 96.2  & 563       & \cmark \\
    X-CLIP-B/16~\textit{(ECCV'22)}~\cite{xclip}                      & CLIP-400M     & \cmark    & 8         & 1 × 1     & 82.3  & 95.4  & 145       & \cmark \\
    X-CLIP-B/16~\textit{(ECCV'22)}~\cite{xclip}                      & CLIP-400M     & \cmark    & 8         & 4 × 3     & 83.8  & 96.7  & 145       & \cmark \\
    X-CLIP-B/16~\textit{(ECCV'22)}~\cite{xclip}                      & CLIP-400M     & \cmark    & 16        & 4 × 3     & 84.7  & 96.8  & 287       & \cmark \\
    \midrule
    \multicolumn{7}{l}{\textit{Initialization: Large-scale image-language weights (frozen backbone)}} \\
    EVL B/16~\textit{(ECCV'22)}~\cite{frozen-clip}                   & CLIP-400M     & \xmark    & 8         & 1 x 3     & \textbf{82.9}  & -     & 444       & \xmark \\
    A6~\textit{(ECCV'22)}~\cite{video-coop}                          & CLIP-400M     & \xmark    & 16        & -         & 76.9  & 93.5  & -         & \cmark \\
    Vita-CLIP B/16 ($M_c = 8, M_v = 8$)                              & CLIP-400M     & \xmark    & 8         & 1 x 1     & 80.5  & 95.9  & 97        & \cmark \\
    Vita-CLIP B/16 ($M_c = 8, M_v = 8$)                              & CLIP-400M     & \xmark    & 8         & 4 x 3     & 81.8  & 96.0  & 97        & \cmark \\
    Vita-CLIP B/16 ($M_c = 8, M_v = 8$)                              & CLIP-400M     & \xmark    & 16        & 4 x 3     & \textbf{82.9}  & \textbf{96.3}  & 190       & \cmark \\
    
    \bottomrule
\end{tabular}}
\label{tab:sup_k400}
\vspace{-.4cm}

\end{table*}

\begin{table}[t]
% \vspace{-.4cm}

\caption{Comparison with supervised methods on Something-Something-V2 \cite{goyal2017ssv2}, with a mention of their zero-shot capability.}
\centering
\small
\setlength{\tabcolsep}{1.0mm}{
\resizebox{0.85\columnwidth}{!}{
\begin{tabular}{lcc}
\rowcolor{Gray}
\toprule
    Method                                  & Zero-shot   & Top-1 \\
    \midrule
    \multicolumn{2}{l}{\textit{Methods with Finetuned Backbone}} \\
    TRN~\textit{(ECCV'18)}~\cite{zhou2018temporal}             & \xmark                & 48.8 \\ 
    SlowFast~\textit{(CVPR'20)}~\cite{feichtenhofer2020x3d}    & \xmark                & 61.7 \\ 
    TSM~\textit{(ICCV'19)}~\cite{lin2019tsm}                   & \xmark                & 63.4 \\ 
    ViViT~\textit{(ICCV'21)}~\cite{arnab2021vivit}             & \xmark                & 65.9 \\ 
    Swin-B~\textit{(CVPR'22)}~\cite{liu2021video}              & \xmark                & 69.6 \\
    \midrule
    \multicolumn{2}{l}{\textit{Methods with Frozen Backbone}} \\
    B2~\textit{(ECCV'22)}~\cite{video-coop}                    & \cmark                & 38.1  \\
    %\midrule
    Vita-CLIP B/16 ($M_c = 8, M_v = 8$)     & \cmark                & \textbf{48.7} \\
    \bottomrule
\end{tabular}
}
}

\label{tab:sup_ssv2}
\hfill
\vspace{-.1cm}
\end{table}

\subsection{Experimental Setup and Protocols}
\label{sec:exp:data}

\noindent\textbf{Datasets:} In the supervised setting, we train on the train set of Kinetics-400 (K400) \cite{k400} and Something-Something-V2 (SSv2) \cite{goyal2017ssv2}). We report supervised performance against existing methods in the literature on the validation sets of K400 and SSv2. For zero-shot experiments, we train on K400 training set and evaluate on three datasets: Kinetics-600 (K600) \cite{k600}, HMDB51 \cite{kuehne2011hmdb} and UCF101 \cite{soomro2012ucf101}. 
For zero-shot evaluation on K600, we follow \cite{chen2021elaborative}, using the 220 new categories outside of (K400) for evaluation. 
Following \cite{xclip}, we conduct evaluation three times, each time randomly sampling 160 categories for evaluation from the 220 categories in (K600).
For zero-shot evaluation on HMDB51 and UCF101, we follow \cite{zoph2018nasnet} and report average top-1 accuracy and standard deviation on three splits of the test set.

\noindent\textbf{Hyperparameters:} For all experiments we train the model for 30 epochs with a cosine decay scheduler and an initial learning rate of $8 \times 10^{-4}$. Unless stated otherwise, the number of frames during training is set to $8$. For evaluation, we use a single view of $8$ frames in a supervised setting. During the zero-shot evaluation, we train the model with $8$ frames but evaluate with a single view of $32$ frames.

\subsection{Supervised Experiments}
\label{sec:exp:sup}
In the supervised setting, we present results on K400 and SSv2 in \autoref{tab:sup_k400} and \autoref{tab:sup_ssv2} respectively. We compare against existing methods under various initializations (random, ImageNet-1k/21k \cite{deng2009imagenet} and CLIP400M) and in terms of GFLOPs, training frames and evaluation views. %To ensure a fair comparison, where possible, we report results with single view evaluation, unless they are not available in the respective publication.

Comparing Vita-CLIP with the ImageNet pretrained methods, we see that our models perform better or competitively against all others while maintaining much lower GFLOP counts and keeping the entire backbone frozen. We perform better than both TimeSformer \cite{timesformer2021} and Mformer \cite{patrick2021keeping} while having $6\times$ and $4\times$ lower GFLOPs, respectively. We perform on par with Swin-B \cite{liu2021video} (IN-1k) while maintaining competitive results against Swin-B (IN-21k) and MViTv2-B with $2$-$3\times$ lower GFLOPs. Note that each of these models has been fully trained, while our Vita-CLIP only trains the proposed prompting scheme. 

Similarly, comparing Vita-CLIP with CLIP-400M pretrained methods, we achieve $3.6\%$ better top-1 accuracy against the A6 \cite{video-coop} prompting method which also uses a frozen backbone similar to ours. We also perform competitively against both X-CLIP \cite{xclip} and ActionCLIP \cite{action-clip}, both of which fine-tune the pretrained backbone while maintaining a lower GFLOP count. Compared with EVL \cite{frozen-clip}, which also uses a frozen backbone, our performance is save, and we additionally hold two advantages. Firstly, we have $4.5\times$ lower GFLOPs, and secondly, we retain the zero-shot capability while EVL cannot be used for zero-shot recognition.

On SSv2, we compare supervised performance against recent methods in \autoref{tab:sup_ssv2}. While we are lower than cross-entropy-based methods, we surpass the best vision-text-based method B6 \cite{video-coop}, by more than $10\%$. Note that the performance for vision-language models is consistently lower than cross-entropy ones. This is due to the fine-grained nature of the SSv2 class descriptions, which are more difficult to differentiate compared to, for example, K400 classes.

From the above experiments, we can see that our Vita-CLIP performs better or competitive against existing methods while maintaining the capability of zero-shot inference. This can be attributed to our prompting scheme that helps capture both the per-frame variation (through the local frame-level prompts) as well as the overall distribution of the video and the dataset (through the summary token and the global video-level prompts respectively).

% tables: zero-shot
\begin{table}[t!]
% \vspace{-.4cm}

\caption{Comparison for zero-shot performances on HMDB51 \cite{kuehne2011hmdb} and UCF101 \cite{soomro2012ucf101} against state-of-the-art.}
\centering
\small
\setlength{\tabcolsep}{1.0mm}{
\resizebox{0.95\columnwidth}{!}{
\begin{tabular}{lcc}
\rowcolor{Gray}
    \toprule
	Method                                                    & HMDB-51           & UCF-101 \\
    \midrule
    \multicolumn{2}{l}{\textit{Methods with Vision Training}} \\
	ASR~\textit{(ECML'17)}~\cite{wang2017alternative}         & 21.8 $\pm$ 0.9    & 24.4 $\pm$ 1.0 \\
	ZSECOC~\textit{(CVPR'17)}~\cite{qin2017zero}              &  22.6 $\pm$ 1.2   & 15.1 $\pm$ 1.7 \\
	UR~\textit{(CVPR'18)}~\cite{zhu2018towards}               & 24.4 $\pm$ 1.6    & 17.5 $\pm$ 1.6 \\
	TS-GCN~\textit{(AAAI'19)}~\cite{gao2019know}              & 23.2 $\pm$ 3.0    & 34.2 $\pm$ 3.1 \\
	E2E~\textit{(CVPR'20)}~\cite{brattoli2020rethinking}      & 32.7              & 48 \\
	ER-ZSAR~\textit{(ICCV'21)}~\cite{chen2021elaborative}     & 35.3 $\pm$ 4.6    & 51.8 $\pm$ 2.9 \\
    \midrule
    \multicolumn{2}{l}{\textit{Methods with Vision-Language Training}} \\
	ActionCLIP~\textit{(arXiv'21)}~\cite{action-clip}         & 40.8 $\pm$ 5.4    & 58.3 $\pm$ 3.4 \\
    A5~\textit{(ECCV'22)}~\cite{video-coop}                   & 44.3 $\pm$ 2.2    & 69.3 $\pm$ 4.2 \\
	X-CLIP-B/16~\textit{(ECCV'22)}~\cite{xclip}               & 44.6 $\pm$ 5.2    & 72.0 $\pm$ 2.3 \\
	  Vita-CLIP B/16 ($M_c = 8, M_v = 8$)    & \textbf{48.6  $\pm$ 0.6} & \textbf{75.0 $\pm$ 0.6} \\
    \bottomrule
\end{tabular}
}
}

\label{tab:zeroshot_hmdb_ucf}
\hfill

\end{table}
\begin{table}[t!]
% \vspace{-.4cm}

\caption{Comparison against state-of-the-art on Kinetics-600 \cite{k600} zero-shot performance.}
\centering
\small
\setlength{\tabcolsep}{3.0mm}{
\resizebox{0.85\columnwidth}{!}{
\begin{tabular}{lcc}
\rowcolor{Gray}
    \toprule
	Method                                  & Top-1 \\
    \midrule
    \multicolumn{2}{l}{\textit{Methods with Vision Training}}\\
    SJE~\textit{(ICCV'15)}~\cite{akata2015evaluation}          & 22.3 $\pm$ 0.6 \\
    ESZSL~\textit{(ICML'15)}~\cite{romera2015embarrassingly}   & 22.9 $\pm$ 1.2 \\
    DEM~\textit{(CVPR'17)}~\cite{zhang2017learning}            & 23.6 $\pm$ 0.7 \\
    GCN~\textit{(arXiv'2020)}~\cite{ghosh2020all}                 & 22.3 $\pm$ 0.6 \\  
    ER-ZSAR~\textit{(ICCV'2021)}~\cite{chen2021elaborative}      & 42.1 $\pm$ 1.4 \\
    \midrule
    \multicolumn{2}{l}{\textit{Methods with Vision-Language Training}} \\
    X-CLIP-B/16~\textit{(ECCV'2022)}~\cite{xclip}                & 65.2 $\pm$ 0.4 \\
	  Vita-CLIP B/16 ($M_c = 8, M_v = 8$)     & \textbf{67.4 $\pm$ 0.5} \\
    \bottomrule
\end{tabular}
}
}
\label{tab:zeroshot_k600}
\hfill
\vspace{-.1cm}
\end{table}

\subsection{Zero-shot Experiments}
\label{sec:exp:zs}

As stated earlier, in the zero-shot experiments we train our Vita-CLIP on the K400 training set with $8$ frames, then perform the zero-shot evaluation on three datasets, UCF101 \cite{soomro2012ucf101}, HMDB51 \cite{kuehne2011hmdb} and K600 \cite{k600}. Notably, we utilize the \emph{same model and hyperparameters} as used for the supervised experiments, unlike the current SoTA method X-CLIP \cite{xclip} which uses a different train setting for zero-shot evaluation.

For the zero-shot setting, we simply replace the class-specific context with a tokenized class description. Our results for zero-shot performance on UCF101, HMDB51, and K600 are presented in \autoref{tab:zeroshot_hmdb_ucf} and \autoref{tab:zeroshot_k600} respectively.  
It can be seen from \autoref{tab:zeroshot_hmdb_ucf} that we outperform the previous methods by $4\%$ and $3\%$ respectively on HMDB51 and UCF101. Similarly, we achieve state-of-the-art zero-shot performance on K600, surpassing the previous best by $2.2\%$. We attribute this strong performance to both our proposed prompting scheme, as well as the fact that we retain the pretrained general representation of the CLIP backbone.

\subsection{Supervised vs. Zero-shot Trade-off}
\label{sec:exp:tradeoff}

In this section, we further highlight the trade-off that we attempt to balance through our proposed method. Consider \autoref{tab:tradeoff} where the current state-of-the-art approach X-CLIP \cite{xclip} has two different sets of hyper-parameters for supervised and zero-shot settings. The authors use $8$ frame sampling and train for $30$ epochs in the supervised setting. While in the zero-shot setting, X-CLIP trains for $10$ epochs while sampling $32$ frames per clip. This results in two models which only perform well in either supervised or zero-shot settings, but not both. Instead, our Vita-CLIP, which aims at retaining the generalized representation of the backbone while adapting to videos using prompt learning, is able to achieve a balance between both settings. This allows us to use a single model, trained with sampling $8$ frames per clip, for a total of $30$ epochs to be used in both settings.

% tables: 
\begin{table}[t!]
% \vspace{-.4cm}

\caption{Comparing performance (supervised/zero-shot) and trainable parameter trade-off between X-CLIP \cite{xclip} and Vita-CLIP. (*) indicates results 
%not reported by the authors but 
obtained by the official repository of \cite{xclip}.}
\centering

\setlength{\tabcolsep}{1.0mm}{
\resizebox{0.9\columnwidth}{!}{
\begin{tabular}{lcccc}
\rowcolor{Gray}
\toprule
	Method & \thead{K400 \\ Top 1 \\ Supervised} & \thead{HMDB51 \\ Top 1 \\ Zeroshot} & \thead{UCF101 \\ Top 1 \\ Zeroshot} & \thead{Trainable \\ Parameters \\ (M)} \\
    \midrule
	X-CLIP B/16 (Supervised) & 82.3 & 41.4* & 67.9*   & 131.5 \\
    X-CLIP B/16 (Zero-shot)  & 78.2* & 44.6 & 72.0     & 131.5 \\
	Ours B/16  & 80.5 & 48.6 & 75.0    & 38.88 \\
    \bottomrule
\end{tabular}
}
}
\label{tab:tradeoff}
\hfill
\end{table}

\begin{figure*}[ht]
\centering
        \begin{subfigure}[t]{0.3\textwidth}
        \centering
        \includegraphics[width=\linewidth]{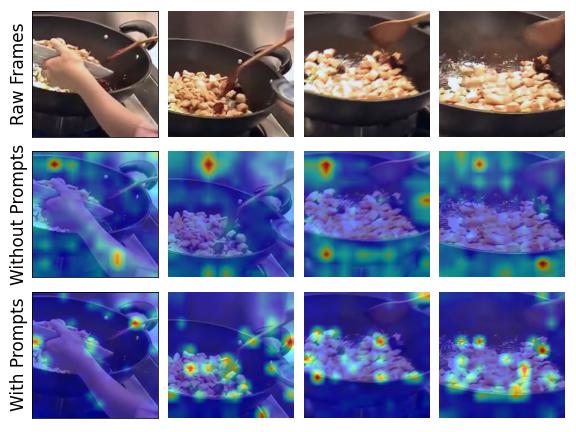}
        \caption{Class Label: "Cooking Chicken"}
        \end{subfigure}
        \begin{subfigure}[t]{0.3\textwidth}
        \centering
        \includegraphics[width=\linewidth]{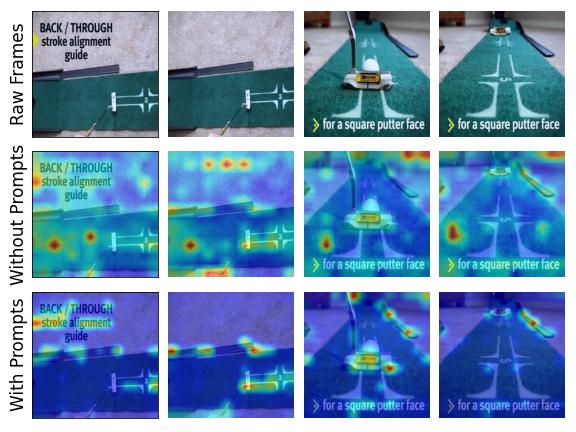}
        \caption{Class Label: "Golf Putting"}
        \end{subfigure}
        \begin{subfigure}[t]{0.3\textwidth}
        \centering
        \includegraphics[width=\linewidth]{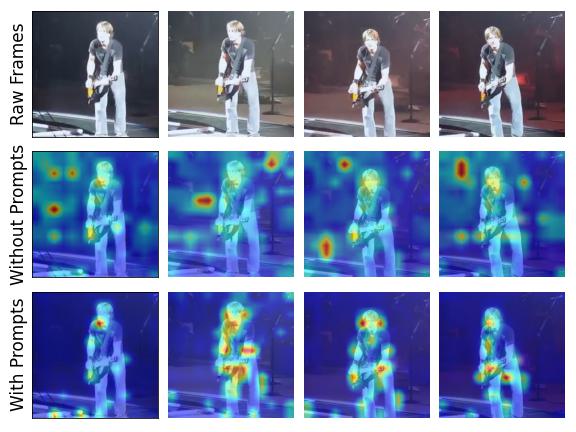}
        \caption{Class Label: "Playing Guitar"}
        \end{subfigure}
    
        \begin{subfigure}[t]{0.3\textwidth}
        \centering
        \includegraphics[width=\linewidth]{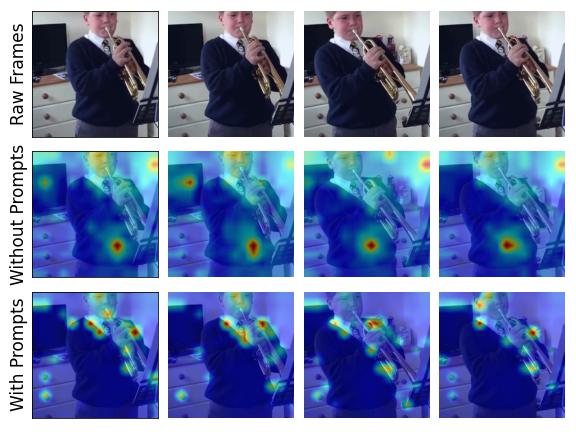}
        \caption{Class Label: "Playing Trumpet"}
        \end{subfigure}
        \begin{subfigure}[t]{0.3\textwidth}
        \centering
        \includegraphics[width=\linewidth]{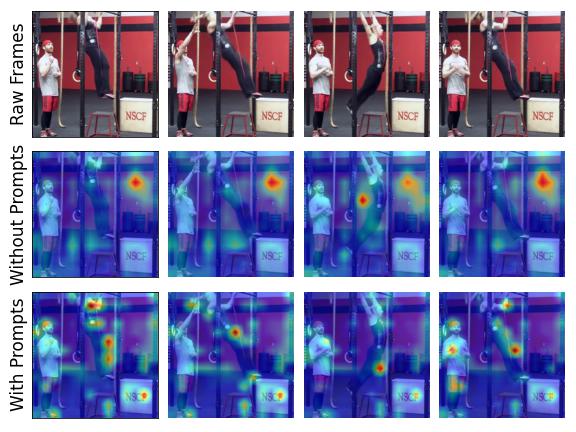}
        \caption{Class Label: "Pull Ups"}
        \end{subfigure}
        \begin{subfigure}[t]{0.3\textwidth}
        \centering
        \includegraphics[width=\linewidth]{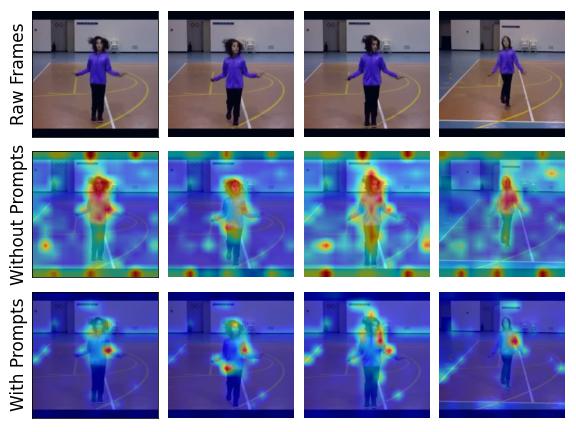}
        \caption{Class Label: "Skipping Rope"}
        \end{subfigure}
    \caption{Attention Rollout \cite{attn_roll} on sample videos showing raw frames, heatmap without our proposed prompting method, and heatmap with Vita-CLIP prompting method. For example, in actions like `Cooking Chicken', `Playing Guitar', `Pull Ups', and `Skipping Rope', our approach fixates on the important localized parts that matter the most in terms of discriminative information and motion properties.}

    \label{fig:vis}
\vspace{-.5cm}
\end{figure*}

\subsection{Ablations}
\label{sec:exp:ablation}

In this section, we present an ablative study on different components of our method. All experiments are performed with training on the K400 training set and testing on the validation set. All models are trained for $30$ epochs, as stated earlier with $8$ frames sampled per video clip.

% tables: ablations
\begin{table}[t!]
% \vspace{-.4cm}

\caption{Ablations for different types of video prompts proposed in this work: Summary Token ($S$), Global Prompts ($G$) and Local Prompts ($L$). Text side prompting is fixed to Class-Specific Context (CSC) with $M_c = 8$ for this ablation.}
\centering

\small
\setlength{\tabcolsep}{2.0mm}{
\resizebox{1\columnwidth}{!}{
\begin{tabular}{lc}
\rowcolor{Gray}
\toprule
	Method             & Top-1 \\
    \midrule
    CLIP B/16 (Zero-shot) & 40.10 \\
    Vita-CLIP B/16 + CSC ($M_c = 8$)     & 73.00 \\
	Vita-CLIP B/16 + CSC ($M_c = 8$) + $G$ ($M_v = 8$)      & 77.83 \\
	Vita-CLIP B/16 + CSC ($M_c = 8$) + $G$ ($M_v = 8$) + $L$    & 79.16 \\
	Vita-CLIP B/16 + CSC ($M_c = 8$) + $G$ ($M_v = 8$) + $L$ + $S$    & 80.51 \\
    \bottomrule
\end{tabular}
}
}
\label{tab:ablation_video_prompts}
\hfill
\vspace{-.1cm}
\end{table}

\textbf{Video Prompting:} We first perform an ablation on the vision side prompting in \autoref{tab:ablation_video_prompts}. Note for this ablation, text-side prompting in all Vita-CLIP models is fixed at $M_c = 8$ using Class-Specific Context. We define a simple baseline, the zero-shot accuracy of the vanilla CLIP \cite{clip}. Building on that we add the Global video-level prompts, $G$ ($M_v = 8$),  while keeping the rest of the model frozen. This achieves $77.83\%$ top 1 accuracy on K400. We then add the Local frame-level prompts ($L$) which push the model to $79.16\%$. The inclusion of summary token brings us up to $80.51\%$. This shows that the three prompting techniques are complementary and contribute to the overall accuracy of the model.

\begin{SCfigure}
\centering
    \centering\resizebox{0.5\columnwidth}{!}{
    % This file was created with tikzplotlib v0.10.1.
\begin{tikzpicture}

\definecolor{darkgray176}{RGB}{176,176,176}
\definecolor{steelblue31119180}{RGB}{31,119,180}

\begin{axis}[
tick align=outside,
tick pos=left,
x grid style={darkgray176},
xlabel={Number of Global Prompts \(\displaystyle M_v\)},
xmajorgrids,
xmin=0, xmax=30,
xtick style={color=black},
y grid style={darkgray176},
ylabel={Top-1 K400 Accuracy},
ymajorgrids,
ymin=79, ymax=82,
ytick style={color=black}
]
\addplot [semithick, steelblue31119180, dashed, mark=*, mark size=3, mark options={solid}]
table {%
5 79.8
8 80.51
10 80.62
15 80.71
25 80.92
};
\end{axis}

\end{tikzpicture}}
    \caption{Ablations for number of Global video-level prompts ($\mathbf{G}^{(l)} = [\mathbf{g}_{1}^{(l)}, \cdots, \mathbf{g}_{M_v}^{(l)}]$) on K400 dataset. The video-side prompting includes local frame-level prompting ($L$) and summary token ($S$), while the text side prompting is fixed to Class-Specific Context (CSC) with $M_c {=} 8$.}
    \label{fig:ablation_num_global}
\end{SCfigure}
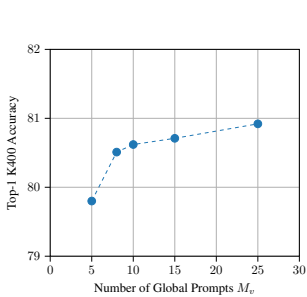

\textbf{Number of Global Video-Level Prompts:} We next evaluate the impact of increasing the number of Global video-level prompts. We test different values for the number of prompts as presented in \autoref{fig:ablation_num_global}. We can see that the accuracy saturates around $M_v=8$, which is why it's the default number of Global prompts we use in all experiments.

\textbf{Number and Type of Text Prompts}: Here, we consider the text-side prompting. We use a baseline where only the tokenized class name is used as context and evaluate two design choices: the number of text prompts $M_c$, and the type of text prompt, Unified Context (UC) (\ie a single set of prompts for all classes), and Class-Specific Context (CSC) (\ie an independent prompt set for each class). The ablation is shown in \autoref{fig:ablation_text}. It is clear that CSC gives better accuracy, which is intuitive given that there is an independent learnable context for each class. Increasing the context size beyond $8$ does not give any significant gain. Thus, we chose to fix the text side prompting to CSC with $M_c = 8$.

\definecolor{steelblue31119180}{RGB}{31,119,180}
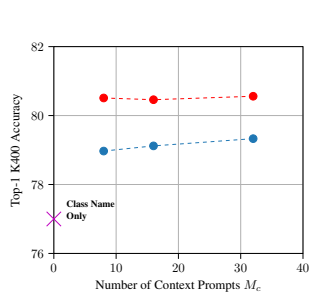
\begin{SCfigure}
\centering
    \centering\resizebox{0.50\columnwidth}{!}{
    % This file was created with tikzplotlib v0.10.1.
\begin{tikzpicture}

\definecolor{darkgray176}{RGB}{176,176,176}
\definecolor{darkviolet1910191}{RGB}{191,0,191}
\definecolor{steelblue31119180}{RGB}{31,119,180}

\begin{axis}[
tick align=outside,
tick pos=left,
x grid style={darkgray176},
xlabel={Number of Context Prompts \(\displaystyle M_c\)},
xmajorgrids,
xmin=0, xmax=40,
xtick style={color=black},
y grid style={darkgray176},
ylabel={Top-1 K400 Accuracy},
ymajorgrids,
ymin=76, ymax=82,
ytick style={color=black}
]
\addplot [semithick, steelblue31119180, dashed, mark=*, mark size=3, mark options={solid}]
table {%
8 78.97
16 79.12
32 79.33
};
\addplot [semithick, red, dashed, mark=*, mark size=3, mark options={solid}]
table {%
8 80.51
16 80.46
32 80.56
};
\draw (axis cs:1.5,77) node[
  scale=0.75,
  anchor=base west,
  text=black,
  rotate=0.0,
  align=left
]{\textbf{Class Name}\\
\textbf{Only}};
\addplot [draw=darkviolet1910191, draw=none, fill=darkviolet1910191, mark=x, mark size=8pt]
table{%
x  y
0 77
};
\end{axis}

\end{tikzpicture}}
    \caption{Ablations for the number of text prompts ($M_c = 8/16/32$) and type of text prompts: \textcolor{steelblue31119180}{Unified Context (UC)} vs. \textcolor{red}{Class-Specific Context (CSC)} on K400. Vision prompting is fixed to global video-level prompting $G$ ($M_v = 8$) with local frame-level prompting $L$ and summary token $S$.}
    \label{fig:ablation_text}
\end{SCfigure}

\textbf{Visualization:} We illustrate the attentions of our model using the attention roll-out \cite{attn_roll} method in \autoref{fig:vis}. We compare the visualizations of our method with a baseline that does not include our proposed prompting scheme. We note that the proposed prompting scheme helps the model to focus on the salient parts and essential dynamics of the video which are relevant to the end recognition task.

\section{Conclusion}\vspace{-0.5em}
We propose a multimodal prompting scheme to adopt image-language pretrained models to the task of video recognition. Existing solutions do not leverage video-text joint prompt learning and often resort to finetuning the CLIP backbone which lacks the balance between zero-shot generalization and supervised performance. Our approach strikes a balance between zero-shot and supervised performance, presenting a unified method that performs well in both settings using the same training scheme. We achieve state-of-the-art zero-shot performance on three datasets (UCF101, HMDB51, and K600) and still remain competitive with respect to supervised performance on K400 and SSv2 while training a much lower number of parameters.

%%%%%%%%% REFERENCES
{\small
\bibliographystyle{ieee_fullname}
\bibliography{egbib}

\begin{thebibliography}{10}\itemsep=-1pt

\bibitem{attn_roll}
Samira Abnar and Willem Zuidema.
\newblock Quantifying attention flow in transformers.
\newblock In {\em Proceedings of the 58th Annual Meeting of the Association for
  Computational Linguistics}, pages 4190--4197, Online, July 2020. Association
  for Computational Linguistics.

\bibitem{akata2015evaluation}
Zeynep Akata, Scott Reed, Daniel Walter, Honglak Lee, and Bernt Schiele.
\newblock Evaluation of output embeddings for fine-grained image
  classification.
\newblock In {\em CVPR}, pages 2927--2936, 2015.

\bibitem{anne2017localizing}
Lisa Anne~Hendricks, Oliver Wang, Eli Shechtman, Josef Sivic, Trevor Darrell,
  and Bryan Russell.
\newblock Localizing moments in video with natural language.
\newblock In {\em ICCV}, pages 5803--5812, 2017.

\bibitem{arnab2021vivit}
Anurag Arnab, Mostafa Dehghani, Georg Heigold, Chen Sun, Mario Lu{\v{c}}i{\'c},
  and Cordelia Schmid.
\newblock Vivit: A video vision transformer.
\newblock In {\em ICCV}, pages 6836--6846, 2021.

\bibitem{bahng2022visual}
Hyojin Bahng, Ali Jahanian, Swami Sankaranarayanan, and Phillip Isola.
\newblock Visual prompting: Modifying pixel space to adapt pre-trained models.
\newblock {\em arXiv preprint arXiv:2203.17274}, 2022.

\bibitem{timesformer2021}
Gedas Bertasius, Heng Wang, and Lorenzo Torresani.
\newblock Is space-time attention all you need for video understanding?
\newblock In {\em ICML}, pages 813--824, 2021.

\bibitem{bertasius2021space}
Gedas Bertasius, Heng Wang, and Lorenzo Torresani.
\newblock Is space-time attention all you need for video understanding?
\newblock In {\em ICML}, July 2021.

\bibitem{brattoli2020rethinking}
Biagio Brattoli, Joseph Tighe, Fedor Zhdanov, Pietro Perona, and Krzysztof
  Chalupka.
\newblock Rethinking zero-shot video classification: End-to-end training for
  realistic applications.
\newblock In {\em CVPR}, pages 4613--4623, 2020.

\bibitem{brown2020language}
Tom Brown, Benjamin Mann, Nick Ryder, Melanie Subbiah, Jared~D Kaplan, Prafulla
  Dhariwal, Arvind Neelakantan, Pranav Shyam, Girish Sastry, Amanda Askell,
  et~al.
\newblock Language models are few-shot learners.
\newblock {\em Advances in neural information processing systems},
  33:1877--1901, 2020.

\bibitem{k600}
Joao Carreira, Eric Noland, Andras Banki-Horvath, Chloe Hillier, and Andrew
  Zisserman.
\newblock A short note about kinetics-600.
\newblock {\em arXiv preprint arXiv:1808.01340}, 2018.

\bibitem{carreira2017quo}
Joao Carreira and Andrew Zisserman.
\newblock Quo vadis, action recognition? a new model and the kinetics dataset.
\newblock In {\em CVPR}, pages 6299--6308, 2017.

\bibitem{chen2021elaborative}
Shizhe Chen and Dong Huang.
\newblock Elaborative rehearsal for zero-shot action recognition.
\newblock In {\em ICCV}, pages 13638--13647, 2021.

\bibitem{deng2009imagenet}
Jia Deng, Wei Dong, Richard Socher, Li-Jia Li, Kai Li, and Li Fei-Fei.
\newblock Imagenet: A large-scale hierarchical image database.
\newblock In {\em CVPR}, pages 248--255. Ieee, 2009.

\bibitem{bert}
Jacob Devlin, Ming-Wei Chang, Kenton Lee, and Kristina Toutanova.
\newblock Bert: Pre-training of deep bidirectional transformers for language
  understanding.
\newblock {\em arXiv preprint arXiv:1810.04805}, 2018.

\bibitem{diba2018spatio}
Ali Diba, Mohsen Fayyaz, Vivek Sharma, M~Mahdi Arzani, Rahman Yousefzadeh,
  Juergen Gall, and Luc Van~Gool.
\newblock Spatio-temporal channel correlation networks for action
  classification.
\newblock In {\em ECCV}, pages 284--299, 2018.

\bibitem{dollar2005behavior}
Piotr Doll{\'a}r, Vincent Rabaud, Garrison Cottrell, and Serge Belongie.
\newblock Behavior recognition via sparse spatio-temporal features.
\newblock In {\em 2005 IEEE international workshop on visual surveillance and
  performance evaluation of tracking and surveillance}, pages 65--72. IEEE,
  2005.

\bibitem{dosovitskiy2020image}
Alexey Dosovitskiy, Lucas Beyer, Alexander Kolesnikov, Dirk Weissenborn,
  Xiaohua Zhai, Thomas Unterthiner, Mostafa Dehghani, Matthias Minderer, Georg
  Heigold, Sylvain Gelly, et~al.
\newblock An image is worth 16x16 words: Transformers for image recognition at
  scale.
\newblock {\em ICLR}, 2021.

\bibitem{du2022learning}
Yu Du, Fangyun Wei, Zihe Zhang, Miaojing Shi, Yue Gao, and Guoqi Li.
\newblock Learning to prompt for open-vocabulary object detection with
  vision-language model.
\newblock In {\em CVPR}, pages 14084--14093, 2022.

\bibitem{fan2021multiscale}
Haoqi Fan, Bo Xiong, Karttikeya Mangalam, Yanghao Li, Zhicheng Yan, Jitendra
  Malik, and Christoph Feichtenhofer.
\newblock Multiscale vision transformers.
\newblock In {\em ICCV}, pages 6824--6835, 2021.

\bibitem{Fan_2021_ICCV}
Haoqi Fan, Bo Xiong, Karttikeya Mangalam, Yanghao Li, Zhicheng Yan, Jitendra
  Malik, and Christoph Feichtenhofer.
\newblock Multiscale vision transformers.
\newblock In {\em ICCV}, pages 6824--6835, 2021.

\bibitem{feichtenhofer2020x3d}
Christoph Feichtenhofer.
\newblock X3d: Expanding architectures for efficient video recognition.
\newblock In {\em CVPR}, 2020.

\bibitem{feichtenhofer2019slowfast}
Christoph Feichtenhofer, Haoqi Fan, Jitendra Malik, and Kaiming He.
\newblock Slowfast networks for video recognition.
\newblock In {\em ICCV}, pages 6202--6211, 2019.

\bibitem{feichtenhofer2016convolutional}
Christoph Feichtenhofer, Axel Pinz, and Andrew Zisserman.
\newblock Convolutional two-stream network fusion for video action recognition.
\newblock In {\em CVPR}, pages 1933--1941, 2016.

\bibitem{gao2019know}
Junyu Gao, Tianzhu Zhang, and Changsheng Xu.
\newblock I know the relationships: Zero-shot action recognition via two-stream
  graph convolutional networks and knowledge graphs.
\newblock In {\em AAAI}, 2019.

\bibitem{clip-adapter}
Peng Gao, Shijie Geng, Renrui Zhang, Teli Ma, Rongyao Fang, Yongfeng Zhang,
  Hongsheng Li, and Yu Qiao.
\newblock Clip-adapter: Better vision-language models with feature adapters.
\newblock {\em arXiv preprint arXiv:2110.04544}, 2021.

\bibitem{gao2020making}
Tianyu Gao, Adam Fisch, and Danqi Chen.
\newblock Making pre-trained language models better few-shot learners.
\newblock {\em arXiv preprint arXiv:2012.15723}, 2020.

\bibitem{ghiasi2021open}
Golnaz Ghiasi, Xiuye Gu, Yin Cui, and Tsung-Yi Lin.
\newblock Open-vocabulary image segmentation.
\newblock {\em arXiv preprint arXiv:2112.12143}, 2021.

\bibitem{ghosh2020all}
Pallabi Ghosh, Nirat Saini, Larry~S Davis, and Abhinav Shrivastava.
\newblock All about knowledge graphs for actions.
\newblock {\em arXiv preprint arXiv:2008.12432}, 2020.

\bibitem{goyal2017ssv2}
Raghav Goyal, Samira Ebrahimi~Kahou, Vincent Michalski, Joanna Materzynska,
  Susanne Westphal, Heuna Kim, Valentin Haenel, Ingo Fruend, Peter Yianilos,
  Moritz Mueller-Freitag, et~al.
\newblock The" something something" video database for learning and evaluating
  visual common sense.
\newblock In {\em ICCV}, pages 5842--5850, 2017.

\bibitem{alignment}
Tengda Han, Weidi Xie, and Andrew Zisserman.
\newblock Temporal alignment networks for long-term video.
\newblock In {\em CVPR}, 2022.

\bibitem{align}
Chao Jia, Yinfei Yang, Ye Xia, Yi{-}Ting Chen, Zarana Parekh, Hieu Pham,
  Quoc~V. Le, Yun{-}Hsuan Sung, Zhen Li, and Tom Duerig.
\newblock Scaling up visual and vision-language representation learning with
  noisy text supervision.
\newblock {\em CoRR}, abs/2102.05918, 2021.

\bibitem{vpt}
Menglin Jia, Luming Tang, Bor-Chun Chen, Claire Cardie, Serge Belongie, Bharath
  Hariharan, and Ser-Nam Lim.
\newblock Visual prompt tuning.
\newblock In {\em ECCV}, 2022.

\bibitem{jia2022visual}
Menglin Jia, Luming Tang, Bor-Chun Chen, Claire Cardie, Serge Belongie, Bharath
  Hariharan, and Ser-Nam Lim.
\newblock Visual prompt tuning.
\newblock {\em arXiv preprint arXiv:2203.12119}, 2022.

\bibitem{jiang2019stm}
Boyuan Jiang, MengMeng Wang, Weihao Gan, Wei Wu, and Junjie Yan.
\newblock Stm: Spatiotemporal and motion encoding for action recognition.
\newblock In {\em ICCV}, pages 2000--2009, 2019.

\bibitem{jiang2020can}
Zhengbao Jiang, Frank~F Xu, Jun Araki, and Graham Neubig.
\newblock How can we know what language models know?
\newblock {\em Transactions of the Association for Computational Linguistics},
  8:423--438, 2020.

\bibitem{video-coop}
Chen Ju, Tengda Han, Kunhao Zheng, Ya Zhang, and Weidi Xie.
\newblock Prompting visual-language models for efficient video understanding.
\newblock In {\em ECCV}. Springer, 2022.

\bibitem{k400}
Will Kay, Joao Carreira, Karen Simonyan, Brian Zhang, Chloe Hillier, Sudheendra
  Vijayanarasimhan, Fabio Viola, Tim Green, Trevor Back, Paul Natsev, et~al.
\newblock The kinetics human action video dataset.
\newblock {\em arXiv preprint arXiv:1705.06950}, 2017.

\bibitem{klaser2008spatio}
Alexander Klaser, Marcin Marsza{\l}ek, and Cordelia Schmid.
\newblock A spatio-temporal descriptor based on 3d-gradients.
\newblock In {\em BMVC}, pages 275--1. British Machine Vision Association,
  2008.

\bibitem{kuehne2011hmdb}
Hildegard Kuehne, Hueihan Jhuang, Est{\'\i}baliz Garrote, Tomaso Poggio, and
  Thomas Serre.
\newblock Hmdb: a large video database for human motion recognition.
\newblock In {\em ICCV}, pages 2556--2563, 2011.

\bibitem{lecun2015deep}
Yann LeCun, Yoshua Bengio, and Geoffrey Hinton.
\newblock Deep learning.
\newblock {\em nature}, 521(7553):436--444, 2015.

\bibitem{lei2021less}
Jie Lei, Linjie Li, Luowei Zhou, Zhe Gan, Tamara~L Berg, Mohit Bansal, and
  Jingjing Liu.
\newblock Less is more: Clipbert for video-and-language learning via sparse
  sampling.
\newblock In {\em CVPR}, pages 7331--7341, 2021.

\bibitem{lester2021power}
Brian Lester, Rami Al-Rfou, and Noah Constant.
\newblock The power of scale for parameter-efficient prompt tuning.
\newblock {\em arXiv preprint arXiv:2104.08691}, 2021.

\bibitem{li2022uniformer}
Kunchang Li, Yali Wang, Junhao Zhang, Peng Gao, Guanglu Song, Yu Liu, Hongsheng
  Li, and Yu Qiao.
\newblock Uniformer: Unifying convolution and self-attention for visual
  recognition.
\newblock In {\em ICLR}, 2022.

\bibitem{li2021prefix}
Xiang~Lisa Li and Percy Liang.
\newblock Prefix-tuning: Optimizing continuous prompts for generation.
\newblock {\em arXiv preprint arXiv:2101.00190}, 2021.

\bibitem{li2021mvitv2}
Yanghao Li, Chao-Yuan Wu, Haoqi Fan, Karttikeya Mangalam, Bo Xiong, Jitendra
  Malik, and Christoph Feichtenhofer.
\newblock Improved multiscale vision transformers for classification and
  detection.
\newblock In {\em CVPR}, 2022.

\bibitem{lin2019tsm}
Ji Lin, Chuang Gan, and Song Han.
\newblock Tsm: Temporal shift module for efficient video understanding.
\newblock In {\em ICCV}, 2019.

\bibitem{frozen-clip}
Ziyi Lin, Shijie Geng, Renrui Zhang, Peng Gao, Gerard de Melo, Xiaogang Wang,
  Jifeng Dai, Yu Qiao, and Hongsheng Li.
\newblock Frozen clip models are efficient video learners.
\newblock {\em arXiv preprint arXiv:2208.03550}, 2022.

\bibitem{liu2021pre}
Pengfei Liu, Weizhe Yuan, Jinlan Fu, Zhengbao Jiang, Hiroaki Hayashi, and
  Graham Neubig.
\newblock Pre-train, prompt, and predict: A systematic survey of prompting
  methods in natural language processing.
\newblock {\em arXiv preprint arXiv:2107.13586}, 2021.

\bibitem{liu2021swin}
Ze Liu, Yutong Lin, Yue Cao, Han Hu, Yixuan Wei, Zheng Zhang, Stephen Lin, and
  Baining Guo.
\newblock Swin transformer: Hierarchical vision transformer using shifted
  windows.
\newblock In {\em ICCV}, pages 10012--10022, 2021.

\bibitem{liu2021video}
Ze Liu, Jia Ning, Yue Cao, Yixuan Wei, Zheng Zhang, Stephen Lin, and Han Hu.
\newblock Video swin transformer.
\newblock In {\em CVPR}, 2022.

\bibitem{clip4clip}
Huaishao Luo, Lei Ji, Ming Zhong, Yang Chen, Wen Lei, Nan Duan, and Tianrui Li.
\newblock {CLIP4Clip}: An empirical study of clip for end to end video clip
  retrieval.
\newblock {\em arXiv preprint arXiv:2104.08860}, 2021.

\bibitem{miech2020end}
Antoine Miech, Jean-Baptiste Alayrac, Lucas Smaira, Ivan Laptev, Josef Sivic,
  and Andrew Zisserman.
\newblock End-to-end learning of visual representations from uncurated
  instructional videos.
\newblock In {\em CVPR}, pages 9879--9889, 2020.

\bibitem{naseer2023boosting}
Muzammal Naseer, Ahmad Mahmood, Salman Khan, and Fahad Khan.
\newblock Boosting adversarial transferability using dynamic cues.
\newblock In {\em ICLR}, 2023.

\bibitem{neimark2021video}
Daniel Neimark, Omri Bar, Maya Zohar, and Dotan Asselmann.
\newblock Video transformer network.
\newblock In {\em ICCV}, pages 3163--3172, 2021.

\bibitem{xclip}
Bolin Ni, Houwen Peng, Minghao Chen, Songyang Zhang, Gaofeng Meng, Jianlong Fu,
  Shiming Xiang, and Haibin Ling.
\newblock Expanding language-image pretrained models for general video
  recognition.
\newblock In {\em ECCV}, 2022.

\bibitem{patrick2021keeping}
Mandela Patrick, Dylan Campbell, Yuki Asano, Ishan Misra, Florian Metze,
  Christoph Feichtenhofer, Andrea Vedaldi, and Jo{\~a}o~F Henriques.
\newblock Keeping your eye on the ball: Trajectory attention in video
  transformers.
\newblock In {\em NeurIPS}, 2021.

\bibitem{qin2017zero}
Jie Qin, Li Liu, Ling Shao, Fumin Shen, Bingbing Ni, Jiaxin Chen, and Yunhong
  Wang.
\newblock Zero-shot action recognition with error-correcting output codes.
\newblock In {\em CVPR}, pages 2833--2842, 2017.

\bibitem{clip}
Alec Radford, Jong~Wook Kim, Chris Hallacy, Aditya Ramesh, Gabriel Goh,
  Sandhini Agarwal, Girish Sastry, Amanda Askell, Pamela Mishkin, Jack Clark,
  Gretchen Krueger, and Ilya Sutskever.
\newblock Learning transferable visual models from natural language
  supervision.
\newblock In Marina Meila and Tong Zhang, editors, {\em ICML}, volume 139 of
  {\em Proceedings of Machine Learning Research}, pages 8748--8763. {PMLR},
  2021.

\bibitem{ranasinghe2022selfsupervised}
Kanchana Ranasinghe, Muzammal Naseer, Salman Khan, Fahad~Shahbaz Khan, and
  Michael Ryoo.
\newblock Self-supervised video transformer.
\newblock In {\em ICCV}, June 2022.

\bibitem{rao2022denseclip}
Yongming Rao, Wenliang Zhao, Guangyi Chen, Yansong Tang, Zheng Zhu, Guan Huang,
  Jie Zhou, and Jiwen Lu.
\newblock Denseclip: Language-guided dense prediction with context-aware
  prompting.
\newblock In {\em CVPR}, pages 18082--18091, 2022.

\bibitem{rohrbach2017movie}
Anna Rohrbach, Atousa Torabi, Marcus Rohrbach, Niket Tandon, Christopher Pal,
  Hugo Larochelle, Aaron Courville, and Bernt Schiele.
\newblock Movie description.
\newblock {\em IJCV}, 123(1):94--120, 2017.

\bibitem{romera2015embarrassingly}
Bernardino Romera-Paredes and Philip Torr.
\newblock An embarrassingly simple approach to zero-shot learning.
\newblock In {\em ICML}, pages 2152--2161, 2015.

\bibitem{schick2020exploiting}
Timo Schick and Hinrich Sch{\"u}tze.
\newblock Exploiting cloze questions for few shot text classification and
  natural language inference.
\newblock {\em arXiv preprint arXiv:2001.07676}, 2020.

\bibitem{soomro2012ucf101}
Khurram Soomro, Amir~Roshan Zamir, and Mubarak Shah.
\newblock Ucf101: A dataset of 101 human actions classes from videos in the
  wild.
\newblock {\em arXiv preprint arXiv:1212.0402}, 2012.

\bibitem{stroud2020d3d}
Jonathan Stroud, David Ross, Chen Sun, Jia Deng, and Rahul Sukthankar.
\newblock D3d: Distilled 3d networks for video action recognition.
\newblock In {\em WACV}, pages 625--634, 2020.

\bibitem{sun2022dualcoop}
Ximeng Sun, Ping Hu, and Kate Saenko.
\newblock Dualcoop: Fast adaptation to multi-label recognition with limited
  annotations.
\newblock {\em arXiv preprint arXiv:2206.09541}, 2022.

\bibitem{tran2018closer}
Du Tran, Heng Wang, Lorenzo Torresani, Jamie Ray, Yann LeCun, and Manohar
  Paluri.
\newblock A closer look at spatiotemporal convolutions for action recognition.
\newblock In {\em CVPR}, pages 6450--6459, 2018.

\bibitem{wang2013dense}
Heng Wang, Alexander Kl{\"a}ser, Cordelia Schmid, and Cheng-Lin Liu.
\newblock Dense trajectories and motion boundary descriptors for action
  recognition.
\newblock {\em IJCV}, 103(1):60--79, 2013.

\bibitem{wang2016temporal}
Limin Wang, Yuanjun Xiong, Zhe Wang, Yu Qiao, Dahua Lin, Xiaoou Tang, and Luc
  Van~Gool.
\newblock Temporal segment networks: Towards good practices for deep action
  recognition.
\newblock In {\em ECCV}, pages 20--36. Springer, 2016.

\bibitem{action-clip}
Mengmeng Wang, Jiazheng Xing, and Yong Liu.
\newblock Actionclip: {A} new paradigm for video action recognition.
\newblock {\em CoRR}, abs/2109.08472, 2021.

\bibitem{wang2017alternative}
Qian Wang and Ke Chen.
\newblock Alternative semantic representations for zero-shot human action
  recognition.
\newblock In {\em ECML PKDD}, pages 87--102, 2017.

\bibitem{wang2017spatiotemporal}
Yunbo Wang, Mingsheng Long, Jianmin Wang, and Philip~S Yu.
\newblock Spatiotemporal pyramid network for video action recognition.
\newblock In {\em CVPR}, pages 1529--1538, 2017.

\bibitem{xie2018rethinking}
Saining Xie, Chen Sun, Jonathan Huang, Zhuowen Tu, and Kevin Murphy.
\newblock Rethinking spatiotemporal feature learning: Speed-accuracy trade-offs
  in video classification.
\newblock In {\em ECCV}, pages 305--321, 2018.

\bibitem{xu2021videoclip}
Hu Xu, Gargi Ghosh, Po-Yao Huang, Dmytro Okhonko, Armen Aghajanyan, Florian
  Metze, Luke Zettlemoyer, and Christoph Feichtenhofer.
\newblock Videoclip: Contrastive pre-training for zero-shot video-text
  understanding.
\newblock {\em arXiv preprint arXiv:2109.14084}, 2021.

\bibitem{florence}
Lu Yuan, Dongdong Chen, Yi{-}Ling Chen, Noel Codella, Xiyang Dai, Jianfeng Gao,
  Houdong Hu, Xuedong Huang, Boxin Li, Chunyuan Li, Ce Liu, Mengchen Liu,
  Zicheng Liu, Yumao Lu, Yu Shi, Lijuan Wang, Jianfeng Wang, Bin Xiao, Zhen
  Xiao, Jianwei Yang, Michael Zeng, Luowei Zhou, and Pengchuan Zhang.
\newblock Florence: {A} new foundation model for computer vision.
\newblock {\em CoRR}, abs/2111.11432, 2021.

\bibitem{zhang2017learning}
Li Zhang, Tao Xiang, and Shaogang Gong.
\newblock Learning a deep embedding model for zero-shot learning.
\newblock In {\em CVPR}, pages 2021--2030, 2017.

\bibitem{tip-adapter}
Renrui Zhang, Rongyao Fang, Peng Gao, Wei Zhang, Kunchang Li, Jifeng Dai, Yu
  Qiao, and Hongsheng Li.
\newblock Tip-adapter: Training-free clip-adapter for better vision-language
  modeling.
\newblock {\em arXiv preprint arXiv:2111.03930}, 2021.

\bibitem{zhang2022pointclip}
Renrui Zhang, Ziyu Guo, Wei Zhang, Kunchang Li, Xupeng Miao, Bin Cui, Yu Qiao,
  Peng Gao, and Hongsheng Li.
\newblock Pointclip: Point cloud understanding by clip.
\newblock In {\em CVPR}, pages 8552--8562, 2022.

\bibitem{zhou2018temporal}
Bolei Zhou, Alex Andonian, Aude Oliva, and Antonio Torralba.
\newblock Temporal relational reasoning in videos.
\newblock In {\em ECCV}, 2018.

\bibitem{zhou2021denseclip}
Chong Zhou, Chen~Change Loy, and Bo Dai.
\newblock Denseclip: Extract free dense labels from clip.
\newblock {\em arXiv preprint arXiv:2112.01071}, 2021.

\bibitem{cocoop}
Kaiyang Zhou, Jingkang Yang, Chen~Change Loy, and Ziwei Liu.
\newblock Conditional prompt learning for vision-language models.
\newblock In {\em CVPR}, 2022.

\bibitem{coop}
Kaiyang Zhou, Jingkang Yang, Chen~Change Loy, and Ziwei Liu.
\newblock Learning to prompt for vision-language models.
\newblock {\em IJCV}, 2022.

\bibitem{zhou2018mict}
Yizhou Zhou, Xiaoyan Sun, Zheng-Jun Zha, and Wenjun Zeng.
\newblock Mict: Mixed 3d/2d convolutional tube for human action recognition.
\newblock In {\em CVPR}, pages 449--458, 2018.

\bibitem{zhu2018towards}
Yi Zhu, Yang Long, Yu Guan, Shawn Newsam, and Ling Shao.
\newblock Towards universal representation for unseen action recognition.
\newblock In {\em CVPR}, pages 9436--9445, 2018.

\bibitem{zoph2018nasnet}
Barret Zoph, Vijay Vasudevan, Jonathon Shlens, and Quoc~V Le.
\newblock Learning transferable architectures for scalable image recognition.
\newblock In {\em CVPR}, pages 8697--8710, 2018.

\end{thebibliography}
}

\end{document}